\documentclass[final]{cvpr}

\usepackage{times}
\usepackage{epsfig}
\usepackage{graphicx}
\usepackage{amsmath}
\usepackage{amssymb}

\usepackage[pagebackref=true,breaklinks=true,colorlinks,bookmarks=false]{hyperref}

\begin{document}

%%%%%%%%% TITLE
\title{Semi-supervised Learning for Dense Object Detection in Retail Scenes}

\author{Jaydeep Chauhan \hspace{12pt} Srikrishna Varadarajan \hspace{12pt} Muktabh Mayank Srivastava \\
\href{https://paralleldots.com}{ParallelDots, Inc.}\\
}
\maketitle

%%%%%%%%% ABSTRACT
\begin{abstract}
Retail scenes usually contain densely packed high number of objects in each image. Standard object detection techniques use fully supervised training methodology. This is highly costly as annotating a large dense retail object detection dataset involves an order of magnitude more effort compared to standard datasets. Hence, we propose semi-supervised learning to effectively use the large amount of unlabeled data available in the retail domain. We adapt a popular self supervised method called noisy student initially proposed for object classification to the task of dense object detection. We show that using unlabeled data with the noisy student training methodology, we can improve the state of the art on precise detection of objects in densely packed retail scenes. We also show that performance of the model increases as you increase the amount of unlabeled data.
\end{abstract}

%%%%%%%%% BODY TEXT
\section{Introduction}
Recent deep learning based object detection algorithms are becoming increasingly popular and are widely used in numerous applications such as healthcare, autonomous vehicles, surveillance etc. due to their breakthrough advancement. In spite of this breakthrough, dense object detection is challenging for many state of the art object detectors. It is a common use-case for some retail companies in detecting densely packed objects in retail stores to maximize sales and store inventory management. In retail store shelves, objects often have the same brand or nearly identical and are in their immediate proximity. The challenge is to precisely estimate the product boundaries overcoming identical patterns.

Lot of progress has been made to accurately detect these objects and first approach is to improve in the network architecture or loss criterion such as Retinanet architecture \cite{focal-loss} incorporating new focal loss, and other methods extending focal loss such as guided anchoring \cite{wang2019region}, ATSS \cite{zhang2019bridging}, Generalized focal loss \cite{li2020generalized}, Verifocal net \cite{zhang2020varifocalnet}. Another simple approach is training with more larger datasets as deep learning based models can solve any problem really well if we have decent amount of labeled data. But the problem here is to manually tag these labels, which is a very time consuming and expensive process, especially when we are dealing with object detection with very high density(usually greater than 100 objects per image). It’s also very difficult to scale this further. But considering a lot of freely available unlabeled data(free grocery images on the internet, frames of a video recording at a store etc), usually their amount is higher than limited labeled dataset. Yet the many applications of deep learning, such as healthcare, only benefit from training on larger datasets,  when the data must be clean and accurately labeled. 

Recently self supervised learning(SSL) is gaining a lot of attention in the machine learning community, which makes use of unlabeled data by generating pseudo-labels and training the model in a supervised manner. This novel method has shown an impressive performance on the variety of tasks in computer vision, especially image classification \cite{spyros2018}, \cite{pierre2018}, \cite{Aaron2018}, \cite{Xinlei2020}, \cite{Byol2020}. In NLP, self supervised learning has been used even before this term phrased. The word2vec\cite{word2vec} popularized this paradigm and the field has rapidly applying this methodology across many diverse problems \cite{bert}, \cite{gpt}, \cite{bart}. It has also been successfully applied to reinforcement learning \cite{ashvin2018}, \cite{curl}, \cite{amyzhang}.

Due to the complexity in dense object detection, pure self supervised learning methods did not give us good results. Hence we resort to a semi-supervised learning using the noisy student training method \cite{noisystuimagenet}. It is a semi-supervised based approach to train our model on a combination of labeled and unlabeled data. Our methodology contains three significant steps :

\begin{enumerate}
    \item Train a teacher model in a purely supervised manner on a small labeled dataset.
    \item Generate pseudo-labels on unlabeled data with the help of teacher network and refine the predictions.
    \item Train a student model with a combination of labeled and generated pseudo-labels. 
\end{enumerate}
We can iterate this process by treating the student model as a teacher to relabel the unlabeled data and training a new student. Using our method we were able to get 0.5\%, 1\% and 1.5\% improvement in mAP on SKU110K dataset while using 1x, 2x and 20x additional unlabeled data respectively.

\begin{figure*}
    \centering
    \includegraphics[width=8cm, height=5cm]{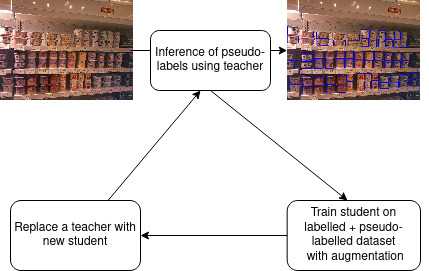} 
    \caption{Self Training Architecture for Object Detection}
    \label{fig:my_label}
\end{figure*}

\section{Related Work}

\textbf{Dense object detection:} In recent years, object detection has been successfully tackled using deep learning \cite{survey}. But the first improvement in dense object detection was made with introduction of focal-loss\cite{focal-loss}. Many further works have been proposed since then, using it as the inspiration, for precisely detecting objects in dense scenes by applying novel techniques and loss functions \cite{li2020generalized}, \cite{zhang2019bridging}, \cite{zhang2020varifocalnet}. Specifically, in the retail domain where dense object detection is more relevant many algorithms have been proposed in the recent years \cite{sku-110k} \cite{sonaal-gaussian} as well.

\textbf{Datasets:} Most of the popular object detection benchmarks such as MS COCO \cite{mscoco}, PASCAL VOC \cite{pascal-voc}, Open Images v4 \cite{open-images}, and ILSVRC \cite{Imagenet} have a very sparse density of objects per image and so might not a valid benchmark for evaluation of the dense object detectors. But recently Goldman\cite{sku-110k} released a benchmark dataset for object detection in a densely crowded scenes. A benchmark of retail datasets was also proposed  \cite{our-benchmark} to increase the diversity and robustness of generic retail product detectors.

\textbf{semi-supervised learning:} Most of the advancement in the field of SSL methods has been limited to image-classification as applying these methods to object detection is hard and costly. For image classification apart from noisy-student \cite{noisystuimagenet}, methods such as consistency regularization have been proposed. They impose regularization on a network to make it robust against the noise and augmentations. Many works are inspired from consistency based methods with some additional components such as mean teacher \cite{mean-teacher}, temporal ensembling \cite{temporal-ensembling}. Fix-match \cite{fixmatch} was proposed combining self training with a consistency loss. Data augmentation is one of the most important aspect of all the SSL methods as they bring drastic improvements. Yet, most of the methods in augmentation are limited to image classification and are unable to translate their success to more complex task of object detection. There some other works that try to apply SSL to the object detection task \cite{STAC}, \cite{CSD} as well.

% \textbf{semi-supervised learning:} Most of the advancement in the field of SSL methods has been limited to image-classification as applying these methods for image classification is comparatively easy and cheaper compared to object detection. For image classification apart from noisy-student \cite{noisystuimagenet}, numerous other diverse methods are available such as consistency regularization, in which by imposing regularization on a network to make it robust against the noise and augmentations. There are a lot of new methods that have been devised from  consistency based methods with some additional components such as mean teacher \cite{mean-teacher}, temporal ensembling \cite{temporal-ensembling} etc. Fix-match \cite{fixmatch} is another impressive method compared to its predecessor mix-match \cite{mixmatch} by combining self training with a consistency loss. Data augmentation is one of the most important pieces of all the SSL methods, which brings drastic improvements. Yet, most of the methods in augmentation are limited to image classification and can't translate their success to object detection because of the complexity, especially in dense object detection, it’s way higher. There some other works that try to apply ssl to object detection field \cite{STAC}, \cite{CSD}.

\begin{figure}
    \centering
    \includegraphics[width=6cm, height=3cm]{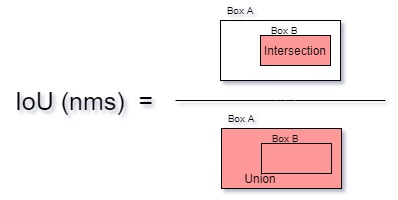}
    \caption{IoU in the case of NMS}
    \label{fig:nms}
\end{figure}

\begin{figure}
    \centering
    \includegraphics[width=8cm, height=3cm]{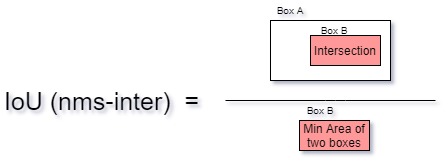}
    \caption{IoU for modified NMS-Inter to remove false positives}
    \label{fig:nms-inter}
\end{figure}

\section{Noisy student training for object detection}

\subsection{Training of teacher network on labeled data}

We use GFL \cite{li2020generalized} as our teacher network for dense object detection as it gets state-of-the-art results on dense object detection in COCO dataset. We trained this model as a teacher on a labeled dataset. GFL (Generalized Focal Loss) extends the focal loss by introducing additional loss terms distributional focal loss, quality focal loss and generalized iou loss. They show that these new losses can significantly improve the performance of object detectors especially on dense object detection. It is able to precisely separate out boundaries of objects of close proximity, and works better in noisy settings compared to previous object detectors.

\begin{table*}[]
    \centering
    \begin{tabular}{|c|c|}

         \hline
         Training Data & SKU110K  \\
         \hline
         PD9k  & 0.389  \\
         PD9k + Pseudo labels (10k) & 0.394  \\
         PD9k + Pseudo labels (20k) & 0.399  \\
         PD9k + Pseudo labels (200k) & 0.403  \\
         \hline

    \end{tabular}
    \caption{Object Detection results on SKU110K dataset using pseudo labels}
    \label{tab:my_label}
\end{table*}

\subsection{Inference of Pseudo-labels on unlabeled data}

After training a teacher network, we did inference on the larger unlabeled data. As the dataset is noisy, we have to perform some heuristics to further refine our predictions. We apply a confidence threshold on the pseudo labels to remove potential false positives while having good recall. This hyper-parameter was estimated by evaluating on a very small amount of data. Non-maximal-suppression was used to refine the predictions further. IoU in calculated as ratio of area of intersection between two boxes and the area of union of the two boxes. We found that using a variation of traditional nms helped better in removing false positives. We call this nms-inter. IoU in the case of nms-inter is the ratio of area of intersection between two boxes and the minimum area of two boxes. This helps in removing false positives predicted completely inside product boxes. 

\section{Datasets}

\subsection{Labeled Dataset}
We use an in-house dataset containing 9000 retail shelf images. We refer to this dataset as PD9k Retail Dataset. This dataset is very diverse consisting of images having various object densities. It consists of both close up images having few SKUs as well as images covering wide shelves having 100-200 SKUs. It contains 926,242 skus and an average density of 102 skus per image.

\subsection{Unlabeled Dataset}
We collected few videos of retail shelves across different shops. We sampled frames from these videos which contain shelves and large number of SKUs using our teacher network. We use these frames as our unlabeled dataset. We refer to this dataset as Retail Video Frames (RVF) Dataset.

\subsection{Test Datasets}
We test the noisy student training on the SKU110K dataset. It is a popular dataset for dense retail object detection. The test split contains a total of 2941 images with 432,312 skus and average density of 147 skus per image.

\section{Experiments}

 We trained our teacher network on PD9k dataset as our baseline model. We further perform three experiments to show the effectiveness of semi-supervised learning. From the RVF Datasets, we randomly sample unlabeled images and created three separate datasets with 10000, 20000 and 200000 images respectively. 
 
 After that, we created a total of three separated datasets which entirely contained tagged dataset that we had used for training teacher network and additional three unlabeled dataset in each dataset as mentioned above. Then we conducted three experiments of noisy student training with respective training dataset. In each experiment, once the training of the student network is complete, we pick the best checkpoint based on mAP on the validation dataset to generate pseudo labels on unlabeled images in the next iteration. 

% (Tbd mention tables and discuss findings)

% \begin{tabular}{ |p{3cm}|p{2cm}|p{2cm}|  }
%  \hline
%  Dataset & Iteration 1 & Iteration 2\\
%  \hline
%  Teacher network(baseline)  & 0.389 & - \\
%  Pseudo labels(10k) & 0.394 & 0.399 \\
%  Pseudo labels(20k) & 0.399 & 0.401 \\
%  Pseudo labels(200K) & 0.403 & - \\
%  \hline
% \end{tabular}

\section{Implementation details}

We used mmdetection framework \cite{mmdet} to train GFL model with Resnet-50 network as a backbone for our dense object detector. For training the teacher network, the network was initialized by pretrained weights based on MS-COCO dataset. All of our training experiments were conducted using SGD optimizer with a learning rate of 1e-3, with momentum 0.9, weight decay 1e-4. We have used step lr scheduler and the model is trained for 30 epochs. For evaluation, pycocotools api was used and mAP was calculated with IOU threshold[0.5:0.95]. We set the maxDet parameter of pycocotools to 300 as used in \cite{sku-110k}.

% (tbd what else can we mention ?)

% After every training epoch, we performed inference on the SKU110K test set with hyperparams score threshold 0.4, nms 0.6 and to maintain our benchmark we have done topk filtering of detection boxes with k = 1000 before applying nms.

\section{Results}
 Our teacher network, which is trained on purely labeled data is the baseline model. It achieves an mAP of 0.389. Table \ref{tab:my_label} summarizes the results for the semi-supervised noisy student model using different amount of unlabeled data. For the model which is trained on 10k pseudo-labeled images with addition to labeled images, we get 0.5\% gain on mAP with respect to baseline. We found 1\% gain in mAP when we trained the model on 20k pseudo-labeled images along with labeled images. In the last model, we take 200k pseudo-labeled images and found 1.4\% gain in mAP compared to baseline model.

% So we compare our experiment results with that baseline.

% For the model which is trained on 10k pseudo-labeled data with addition to labeled data, we get 0.5\% gain for the first iteration and 1\% gain for second iteration of noisy student on mAP with respect to baseline. We found 1\% and 1.1\% gain in first and second iterations when we trained the model on 20k pseudo-labeled data along with labeled data.

% (baseline mAP is 0.389 of the teacher network, which is purely trained on labeled dataset(23k images).

% \section{Conclusion}

{\small
\bibliographystyle{ieee_fullname}
\bibliography{cvpr}
}

\end{document}